%% file: main.tex
\documentclass{article}

    \usepackage[nonatbib,preprint]{neurips_2021}

\usepackage[utf8]{inputenc} 
\usepackage[T1]{fontenc}    
\usepackage{hyperref}       
\usepackage{url}            
\usepackage{booktabs}       
\usepackage{amsfonts}       
\usepackage{nicefrac}       
\usepackage{microtype}      
\usepackage{graphicx}
\usepackage{amsmath}
\usepackage{amssymb}
\usepackage{bm}
\usepackage{multirow}
\usepackage[dvipsnames]{xcolor}

\newcommand{\gf}[1]{{\textbf{\color{Green}{#1}}}} 
\newcommand{\bd}[1]{{\color{Blue}{\underline{#1}}}} 
\newcommand{\etal}{\emph{et~al.}}
\newcommand{\eg}{\textit{e.g.}}

\title{On the Generalization of BasicVSR++\\to Video Deblurring and Denoising\\\vspace{0.3cm}\large{--Technical Report--}}

\author{Kelvin C.K. Chan\qquad Shangchen Zhou\qquad  Xiangyu Xu\qquad Chen Change Loy\vspace{0.05cm}\\
S-Lab, Nanyang Technological University\\
{\tt\small \{chan0899, s200094, xiangyu.xu, ccloy\}@ntu.edu.sg}
}

\begin{document}

\maketitle

\begin{abstract}
    The exploitation of long-term information has been a long-standing problem in video restoration. The recent BasicVSR and BasicVSR++ have shown remarkable performance in video super-resolution through long-term propagation and effective alignment. Their success has led to a question of whether they can be transferred to different video restoration tasks.
    In this work, we extend BasicVSR++ to a generic framework for video restoration tasks. In tasks where inputs and outputs possess identical spatial size, the input resolution is reduced by strided convolutions to maintain efficiency. With only minimal changes from BasicVSR++, the proposed framework achieves compelling performance with great efficiency in various video restoration tasks including video deblurring and denoising. Notably, BasicVSR++ achieves comparable performance to Transformer-based approaches with up to 79\% of parameter reduction and 44${\times}$ speedup. The promising results demonstrate the importance of propagation and alignment in video restoration tasks beyond just video super-resolution. Code and models are available at \url{https://github.com/ckkelvinchan/BasicVSR_PlusPlus}.
\end{abstract}

\input{sections/1_introduction}

\input{sections/2_method}

\input{sections/3_experiments}

\small
\bibliographystyle{ieee_fullname}
\bibliography{short,bib}

\end{document}

%% file: sections/1_introduction.tex
\section{Introduction}
Long-term propagation and effective alignment has been shown essential in video super-resolution~\cite{haris2019recurrent,wang2019edvr,xue2019video}. In our previous work~\cite{chan2021basicvsr}, we summarize existing video super-resolution pipelines into four components, namely \textit{alignment}, \textit{propagation}, \textit{aggregation}, and \textit{upsampling}. Based on the decomposition, \mbox{\textbf{BasicVSR}} is proposed with simple designs. Without dedicated components for video super-resolution, BasicVSR demonstrates superior performance and outperforms state of the arts with improved efficiency. The simplicity and effectiveness of BasicVSR demonstrate the potential of the recurrent framework for video super-resolution.

\begin{figure*}[t]
    \begin{center}
        \includegraphics[width=\textwidth]{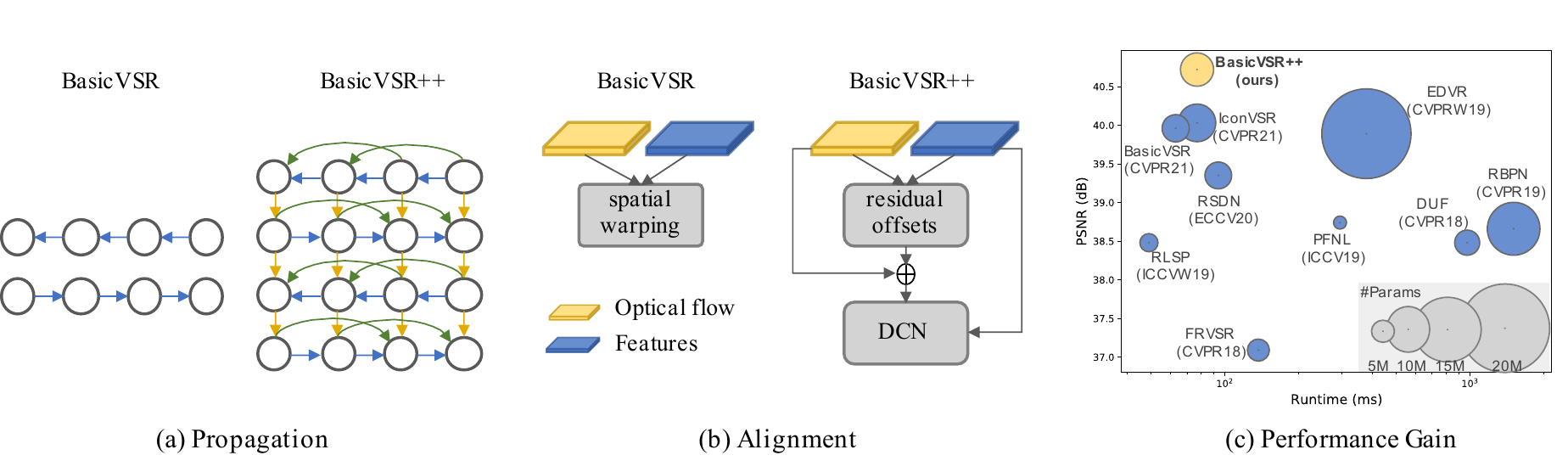}
        \caption{\textbf{Architecture and performance of BasicVSR++.} \textbf{(a)} Second-order grid propagation in BasicVSR++ allows a more effective propagation of features. \textbf{(b)} Flow-guided deformable alignment in BasicVSR++ provides a means for more robust feature alignment across misaligned frames. \textbf{(c)} BasicVSR++ outperforms existing state-of-the-art super-resolution methods including its predecessor, BasicVSR, while maintaining efficiency.}
        \label{fig:teaser}
    \end{center}
\end{figure*}

Motivated by the success, we further improve BasicVSR by replacing the primitive propagation and alignment modules with more sophisticated designs. We propose \mbox{\textbf{BasicVSR++}} with two modifications to employ long-term information more effectively, as shown in Fig.~\ref{fig:teaser}.
First, second-order grid propagation is proposed to more aggressively transmit information across video frames: (1) The second-order connection extends the conventional approach of nearest-frame propagation and distributes information also to the second-next timestamp. In such a way, gradient vanishing can be partially alleviated, and information can be propagated to further timestamps. (2) Grid propagation refines the intermediate features through propagation. Specifically, instead of propagating the features along each direction once, the features are circulated back-and-forth for feature refinement, exploiting long-term information.
Second, BasicVSR++ goes beyond flow-based alignment in BasicVSR and adopts \textit{flow-guided deformable alignment}, combining the motion prior from optical flow and the flexibility of deformable alignment~\cite{chan2021understanding}.
The main idea is to adopt optical flow as the base offsets, and residual offsets are learned. The combined offsets are then used in deformable convolution~\cite{dai2017deformable,zhu2019deformable} for feature alignment. Such a design alleviates the training instability of deformable alignment and improve the alignment accuracy.

As it is a common goal to exploit temporal information in video restoration, we hypothesize that the success of BasicVSR++ is not limited to video super-resolution. In this work, we extend our scope to a more video restoration tasks and introduce a generic framework built on BasicVSR++. For tasks where inputs and outputs possess the same resolution, we reduce the input resolution by strided convolutions to maintain efficiency.
In addition to video super-resolution~\cite{son2021ntire} and compressed video enhancement~\cite{yang2021ntire}, we show that BasicVSR++ is generalizable to video deblurring and denoising, achieving promising performance with high efficiency.

%% file: sections/2_method.tex
\section{Video Restoration Framework}
The original BasicVSR++ assumes that the input resolution is $4{\times}$ smaller than the output resolution. In this work, we extend BasicVSR++ to a generic video restoration framework.
In cases where the input resolution is equal to output resolution (\eg,~deblurring, denoising), we introduce the following two modifications:
\begin{enumerate}
    \item To improve efficiency, we apply strided convolution to the input frames to reduce the spatial resolution.
    \item To further reduce the computational cost, we downsample the input frames to the reduced resolution for optical flow computation.
\end{enumerate}
With these designs, most of the computations are performed in the low-resolution feature space, substantially improving efficiency. The remaining operations follow that of BasicVSR++. We refer readers to the original paper~\cite{chan2021basicvsrpp} for more details.

%% file: sections/3_experiments.tex
\section{Experiments}
In this section, we discuss the performance in \textit{video deblurring} and \textit{video denoising}. For video super-resolution and compressed video enhancement, we refer readers to our original paper~\cite{chan2021basicvsrpp} and the NTIRE~2021 challenge report~\cite{yang2021ntire}, respectively.

\subsection{Video Deblurring}
\noindent\textbf{Settings.}
We mostly follow the settings of the original BasicVSR++, except that we increase the number of residual blocks in each of the four propagation branches from 7 to 15. In addition, strided convolutions are used to reduce the spatial resolution by 2 or 4 times. We adopt Adam optimizer~\cite{kingma2014adam} and Cosine Annealing scheme~\cite{loshchilov2016sgdr}. The initial learning rate of the main network and the flow network are set to $1{\times}10^{-4}$ and $2.5{\times}10^{-5}$, respectively. The weights of the flow network are fixed during the first 5,000 iterations. The batch size is 8 and the patch size of input frames is $256{\times}256$. We use Charbonnier loss~\cite{charbonnier1994two} since it better handles outliers and improves the performance over the conventional $\ell_2$-loss~\cite{lai2017deep}. The number of iterations is set to 600,000 and 200,000 (300,000 for BasicVSR$_{\downarrow2}$) when training on DVD~\cite{su2017deep} and GoPro~\cite{nah2017deep}, respectively. During training, $30$ frames are used as inputs. For testing, we take the full video sequence as inputs to explore information from all video frames for restoration. The detailed configurations can be found in \url{https://github.com/ckkelvinchan/BasicVSR_PlusPlus} and MMEditing~\cite{mmedit}.

\begin{table}[!t]
    \caption{\textbf{Quantitative comparison on DVD~\cite{su2017deep} (Video Deblurring).} \gf{Green} and \bd{blue} colors indicate the best and the second-best performance, respectively.}
    \label{tab:dvd}
    \begin{center}
        \tabcolsep=0.15cm
        \scalebox{0.55}{
            \begin{tabular}{l|c|c|c|c|c|c|c|c|c|c|c}
                     & EDVR~\cite{wang2019edvr} & Tao~\etal~\cite{tao2018scale} & Su~\etal~\cite{su2017deep} & DBLRNet~\cite{zhang2018adversarial} & STFAN~\cite{zhou2019spatiotemporal} & Xiang~\etal~\cite{xiang2020deep} & TSP~\cite{pan2020cascaded} & Suin~\etal~\cite{suin2021gated} & ARVo~\cite{li2021arvo} & \textbf{BasicVSR++$_{\downarrow4}$} & \textbf{BasicVSR++$_{\downarrow2}$} \\\hline
                PSNR & 28.51                    & 29.98                         & 30.01                      & 30.08                               & 31.15                               & 31.68                            & 32.13                      & 32.53                           & 32.80                  & \bd{34.28}                          & \gf{34.78}                          \\
                SSIM & 0.864                    & 0.884                         & 0.888                      & 0.885                               & 0.905                               & 0.916                            & 0.827                      & 0.947                           & 0.935                  & \bd{0.951}                          & \gf{0.956}
            \end{tabular}}
    \end{center}
\end{table}

\begin{table}[!t]
    \caption{\textbf{Quantitative comparison on GoPro~\cite{nah2017deep} (Video Deblurring).} \gf{Green} and \bd{blue} colors indicate the best and the second-best performance, respectively.}
    \label{tab:gopro}
    \begin{center}
        \tabcolsep=0.15cm
        \scalebox{0.55}{
            \begin{tabular}{l|c|c|c|c|c|c|c|c|c|c|c}
                     & RDN~\cite{wieschollek2017learning} & Kim~\etal~\cite{hyun2015generalized} & EDVR~\cite{wang2019edvr} & Su~\etal~\cite{su2017deep} & STFAN~\cite{zhou2019spatiotemporal} & Nah~\etal~\cite{nah2019recurrent} & Tao~\etal~\cite{tao2018scale} & TSP~\cite{pan2020cascaded} & Suin~\etal~\cite{suin2021gated} & \textbf{BasicVSR++$_{\downarrow4}$} & \textbf{BasicVSR++$_{\downarrow2}$} \\\hline
                PSNR & 25.19                              & 26.82                                & 26.83                    & 27.31                      & 28.59                               & 29.97                             & 30.29                         & 31.67                      & 32.10                           & \bd{34.01}                          & \gf{35.08}                          \\
                SSIM & 0.779                              & 0.825                                & 0.843                    & 0.826                      & 0.861                               & 0.895                             & 0.901                         & 0.928                      & \bd{0.960}                      & 0.952                               & \gf{0.961}                          \\
            \end{tabular}}
    \end{center}
\end{table}

\begin{table}[!t]
    \caption{\textbf{Complexity comparison on DVD~\cite{su2017deep} (Video Deblurring).} \gf{Green} and \bd{blue} colors indicate the best and the second-best performance, respectively. Runtime and FLOPs are measured on an RTX~2080~Ti GPU with spatial resolutions $1280{\times}720$ and $240{\times}240$, respectively.}
    \label{tab:complexity}
    \begin{center}
        \tabcolsep=0.15cm
        \scalebox{0.9}{
            \begin{tabular}{l|c|c|c|c|c|c}
                             & EDVR~\cite{wang2019edvr} & Su~\etal~\cite{su2017deep} & STFAN~\cite{zhou2019spatiotemporal} & TSP~\cite{pan2020cascaded} & \textbf{BasicVSR++$_{\downarrow4}$} & \textbf{BasicVSR++$_{\downarrow2}$} \\\hline
                PSNR (dB)    & 28.51                    & 30.01                      & 31.15                               & 32.13                      & \bd{34.28}                          & \gf{34.78}                          \\
                Params (M)   & 23.60                    & 15.30                      & \gf{5.37}                           & 16.19                      & 9.76                                & \bd{9.54}                           \\
                FLOPs (G)    & 159.2                    & 38.7                       & \gf{35.4}                           & 357.9                      & \bd{37.6}                           & 118.0                               \\
                Runtime (ms) & 268.5                    & \bd{133.2}                 & 145.9                               & 579.7                      & \gf{130.5}                          & 433.1
            \end{tabular}}
    \end{center}
\end{table}

\begin{figure*}[!t]
    \begin{center}
        \includegraphics[width=0.95\textwidth]{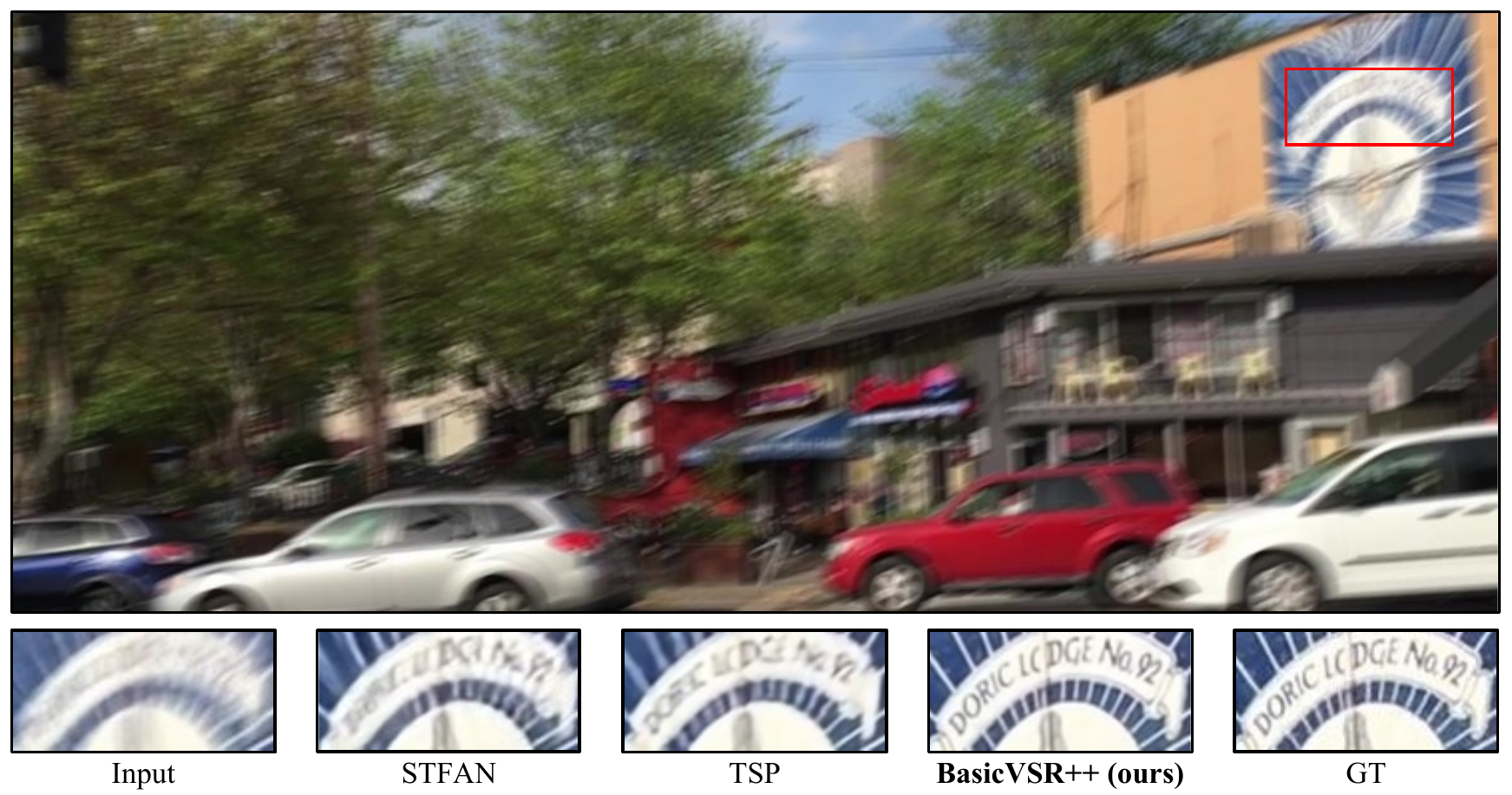}
        \caption{\textbf{Qualitative comparison on DVD~\cite{su2017deep} (Video Deblurring).} Only BasicVSR++$_{\downarrow4}$ is able to restore the the word ``DORIC''.}
        \label{fig:qualitative_deblur}
    \end{center}
\end{figure*}

\noindent\textbf{Quantitative Comparison.}
As shown in Table~\ref{tab:dvd} and Table~\ref{tab:gopro}, BasicVSR++ outperforms existing works by a large margin in terms of PSNR on both DVD~\cite{su2017deep} and GoPro~\cite{nah2017deep}. Notably, BasicVSR++$_{\downarrow4}$ outperforms the second-best methods by \textbf{1.48~dB} and \textbf{1.91~dB} on DVD and GoPro, respectively. Furthermore, thanks to the downsampling mechanism at the input end, it possesses high efficiency. As shown in Table~\ref{tab:complexity}, BasicVSR++$_{\downarrow4}$ has the fastest speed while achieving the highest PSNR. BasicVSR++$_{\downarrow2}$ further improves the performance with \textbf{1.81~dB} and \textbf{2.98~dB} difference on DVD and GoPro, respectively, but with a slower speed.

\noindent\textbf{Qualitative Comparison.}
In Fig.~\ref{fig:qualitative_deblur} we compare BasicVSR++$_{\downarrow4}$ with STFAN~\cite{zhou2019spatiotemporal} and TSP~\cite{pan2020cascaded}. Through aggregating long-term information from the video sequence, BasicVSR++$_{\downarrow4}$ is able to restore the blurry texts, whereas other methods failed to recover the details.

\subsection{Video Denoising}
\noindent\textbf{Settings.}
The settings follow that for deblurring in DVD, except that (1) the number of frames used in training is $25$ and (2) the batch size is reduced to $7$. The detailed configurations can be found in \url{https://github.com/ckkelvinchan/BasicVSR_PlusPlus} and MMEditing~\cite{mmedit}.

\noindent\textbf{Quantitative Comparison.}
From Table~\ref{tab:davis} and Table~\ref{tab:set8}, it is observed that BasicVSR++ outperforms existing works with a much higher efficiency. For example, as shown in Table~\ref{tab:runtime}, BasicVSR++$_{\downarrow2}$ is \textbf{147}${\times}$ faster than PaCNet~\cite{vaksman2021patch} while having 1.97~dB and 1.49~dB improvements on DAVIS~\cite{khoreva2018video} and Set8~\cite{tassano2020fastdvdnet}, respectively. Interestingly, we find that the improvements over previous state of the arts increase with the noise level $\sigma$. We conjecture that long-term information is more important for low-quality videos, where most information is lost due to the severe noise.

\begin{figure*}[t]
    \begin{center}
        \includegraphics[width=0.95\textwidth]{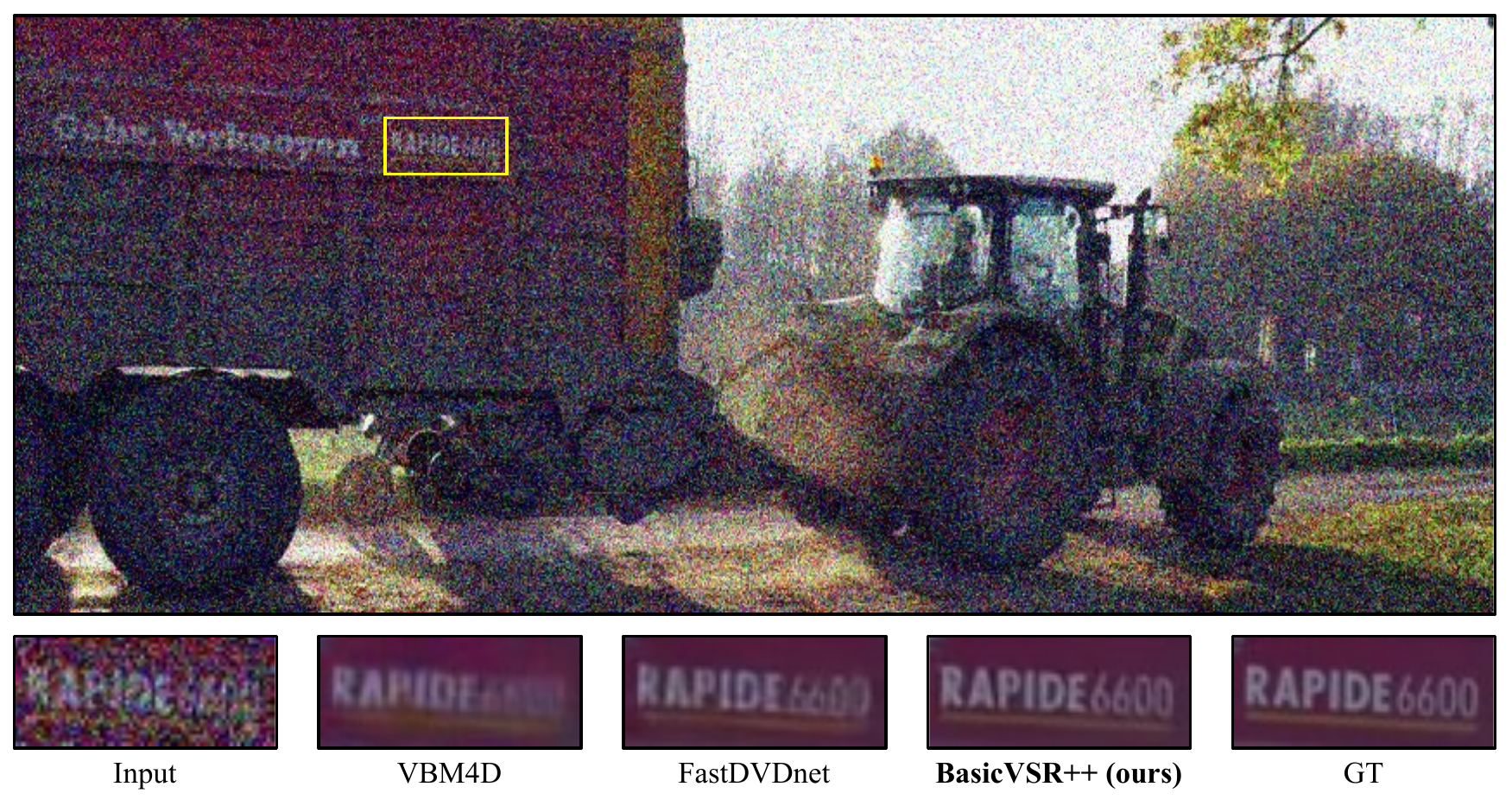}
        \caption{\textbf{Qualitative comparison on Set8~\cite{khoreva2018video} (Video Denoising).} Only BasicVSR++$_{\downarrow4}$ is able to restore the number ``6600''.}
        \label{fig:qualitative_denoise}
    \end{center}
\end{figure*}

\noindent\textbf{Qualitative Comparison.}
In Fig.~\ref{fig:qualitative_denoise} we show an example of $\sigma{=}50$. With severe noise, VBM4D~\cite{maggioni2011video} and FastDVDnet~\cite{tassano2020fastdvdnet} are unable to restore the numbers faithfully. In contrast, the effective propagation and alignment of BasicVSR++$_{\downarrow4}$ lead to a better output with more details revealed.

\begin{table}[!t]
    \caption{\textbf{Quantitative comparison (PSNR/SSIM) on DAVIS~\cite{khoreva2018video} (Video Denoising).} The improvements over existing works increase with the noise level $\sigma$. \gf{Green} and \bd{blue} colors indicate the best and the second-best performance, respectively. $\Delta$ denotes the performance gain over PaCNet~\cite{vaksman2021patch}. Note that PaCNet trains different networks for different noise levels.}
    \vspace{-0.1cm}
    \label{tab:davis}
    \begin{center}
        \tabcolsep=0.15cm
        \scalebox{0.64}{
            \begin{tabular}{l|c|c|c|c|c|c|c|c|c}
                              & VBM4D~\cite{maggioni2011video} & VNLB~\cite{arias2015towards} & DVDnet~\cite{tassano2019dvdnet} & FastDVDnet~\cite{tassano2020fastdvdnet} & VNLNet~\cite{davy2018non} & PaCNet~\cite{vaksman2021patch} & \textbf{BasicVSR++$_{\downarrow4}$} & \textbf{BasicVSR++$_{\downarrow2}$} & $\Delta$    \\\hline
                $\sigma{=}10$ & 37.58/-                        & 38.85/-                      & 38.13/0.9657                    & 38.71/0.9672                            & 39.56/0.9707              & 39.97/0.9713                   & \bd{40.13}/\bd{0.9754}              & \gf{40.97}/\gf{0.9786}              & 1.00/0.0073 \\
                $\sigma{=}20$ & 33.88/-                        & 35.68/-                      & 35.70/0.9422                    & 35.77/0.9405                            & 36.53/0.9464              & 37.10/0.9470                   & \bd{37.41}/\bd{0.9598}              & \gf{38.58}/\gf{0.9666}              & 1.48/0.0196 \\
                $\sigma{=}30$ & 31.65/-                        & 33.73/-                      & 34.08/0.9188                    & 34.04/0.9167                            & -                         & 35.07/0.9211                   & \bd{35.74}/\bd{0.9457}              & \gf{37.14}/\gf{0.9560}              & 2.07/0.0349 \\
                $\sigma{=}40$ & 30.05/-                        & 32.32/-                      & 32.86/0.8962                    & 32.82/0.8949                            & 33.32/0.8996              & 33.57/0.8969                   & \bd{34.49}/\bd{0.9321}              & \gf{36.06}/\gf{0.9459}              & 2.49/0.0490 \\
                $\sigma{=}50$ & 28.80/-                        & 31.13/-                      & 31.85/0.8745                    & 31.86/0.8747                            & -                         & 32.39/0.8743                   & \bd{33.45}/\bd{0.9179}              & \gf{35.18}/\gf{0.9358}              & 2.79/0.0615 \\\hline
                Average       & 32.39/-                        & 34.34/-                      & 34.52/0.9195                    & 31.64/0.9188                            & -                         & 35.62/0.9221                   & \bd{36.24}/\bd{0.9462}              & \gf{37.59}/\gf{0.9566}              & 1.97/0.0345 \\
            \end{tabular}}
    \end{center}
\end{table}

\begin{table}[!t]
    \caption{\textbf{Quantitative comparison (PSNR/SSIM) on Set8~\cite{tassano2020fastdvdnet} (Video Denoising).} The improvements over existing works increase with the noise level $\sigma$. \gf{Green} and \bd{blue} colors indicate the best and the second-best performance, respectively. $\Delta$ denotes the performance gain over PaCNet~\cite{vaksman2021patch}. Note that PaCNet trains different networks for different noise levels.}
    \vspace{-0.1cm}
    \label{tab:set8}
    \begin{center}
        \tabcolsep=0.15cm
        \scalebox{0.64}{
            \begin{tabular}{l|c|c|c|c|c|c|c|c|c}
                              & VBM4D~\cite{maggioni2011video} & VNLB~\cite{arias2015towards} & DVDnet~\cite{tassano2019dvdnet} & FastDVDnet~\cite{tassano2020fastdvdnet} & VNLNet~\cite{davy2018non} & PaCNet~\cite{vaksman2021patch} & \textbf{BasicVSR++$_{\downarrow4}$} & \textbf{BasicVSR++$_{\downarrow2}$} & $\Delta$    \\\hline
                $\sigma{=}10$ & 36.05/-                        & 37.26/-                      & 36.08/0.9510                    & 36.44/0.9540                            & \bd{37.28}/\bd{0.9606}    & 37.06/0.9590                   & 36.83/0.9574                        & \gf{37.67}/\gf{0.9632}              & 0.61/0.0042 \\
                $\sigma{=}20$ & 32.18/-                        & 33.72/-                      & 33.49/0.9182                    & 33.43/0.9196                            & 34.08/0.9273              & 33.94/0.9247                   & \bd{34.15}/\bd{0.9319}              & \gf{35.17}/\gf{0.9415}              & 1.23/0.0168 \\
                $\sigma{=}30$ & 30.00/-                        & 31.74/-                      & 31.68/0.8862                    & 31.68/0.8889                            & -                         & 32.05/0.8921                   & \bd{32.57}/\bd{0.9095}              & \gf{33.68}/\gf{0.9221}              & 1.63/0.0300 \\
                $\sigma{=}40$ & 28.48/-                        & 30.39/-                      & 30.46/0.8564                    & 30.46/0.8608                            & 30.72/0.8622              & 30.70/0.8623                   & \bd{31.42}/\bd{0.8889}              & \gf{32.60}/\gf{0.9043}              & 1.90/0.0420 \\
                $\sigma{=}50$ & 27.33/-                        & 29.24/-                      & 29.53/0.8289                    & 29.53/0.8351                            & -                         & \gf{29.66}/\gf{0.8349}         & \bd{30.49}/\bd{0.8692}              & \gf{31.75}/\gf{0.8878}              & 2.09/0.0529 \\\hline
                Average       & 30.81/-                        & 32.47/-                      & 32.29/0.8881                    & 32.31/0.8917                            & -                         & 32.68/0.8946                   & \bd{33.09}/\bd{0.9114}              & \gf{34.17}/\gf{0.9238}              & 1.49/0.0292
            \end{tabular}}
    \end{center}
\end{table}

\begin{table}[!t]
    \caption{\textbf{Runtime comparison on Set8~\cite{tassano2020fastdvdnet} (Video Denoising).} Notably, BasicVSR++$_{\downarrow2}$ is $147{\times}$ faster than PaCNet with 1.49~dB improvement in PSNR. \gf{Green} and \bd{blue} colors indicate the best and the second-best performance, respectively. Runtime is measured on an RTX~2080~Ti GPU.}
    \label{tab:runtime}
    \begin{center}
        \tabcolsep=0.15cm
        \scalebox{0.69}{
            \begin{tabular}{l|c|c|c|c|c|c|c|c}
                            & VBM4D~\cite{maggioni2011video} & VNLB~\cite{arias2015towards} & DVDnet~\cite{tassano2019dvdnet} & FastDVDnet~\cite{tassano2020fastdvdnet} & VNLNet~\cite{davy2018non} & PaCNet~\cite{vaksman2021patch} & \textbf{BasicVSR++$_{\downarrow4}$} & \textbf{BasicVSR++$_{\downarrow2}$} \\\hline
                PSNR (dB)   & 30.81                          & 32.47                        & 32.29                           & 32.31                                   & -                         & 32.68                          & \bd{33.09}                          & \gf{34.17}                          \\
                Runtime (s) & 420.0                          & 156.0                        & 2.51                            & \gf{0.08}                               & 1.65                      & 35.24                          & \gf{0.08}                           & 0.24
            \end{tabular}}
    \end{center}
\end{table}

\begin{table}[!t]
    \caption{\textbf{Comparison (PSNR~(dB) / Params~(M) / Runtime~(ms)) with Transformer-based methods.} BasicVSR++ achieves comparable performance to Transformer-based methods with better efficiency. \gf{Green} and \bd{blue} colors indicate the best and the second-best performance, respectively. The case of $\sigma{=}50$ is reported for denoising. Runtime is measured on an RTX~2080~Ti GPU with an output resolution of $1280{\times}720$.}
    \label{tab:transformer}
    \begin{center}
        \tabcolsep=0.15cm
        \scalebox{0.8}{
            \begin{tabular}{l|c|c|c|c|c}
                
                                       & VSRT~\cite{cao2021video} & FGST~\cite{lin2022flow} & VRT~\cite{liang2022vrt}      & \textbf{BasicVSR++$_{\downarrow4}$} & \textbf{BasicVSR++$_{\downarrow2}$} \\\hline
                $4{\times}$ SR~(REDS4) & 31.06 / \bd{32.6} / 4312 & -                       & \bd{32.19} / 35.6 / \bd{243} & \gf{32.39} / \gf{7.3} / \gf{98}     & -                                   \\
                Deblurring~(DVD)       & -                        & 33.36 / \bd{9.7} / 247  & 34.27 / 18.3 / \bd{220}      & \bd{34.28} / 9.8 / \gf{131}         & \gf{34.61}/\gf{9.5}/433             \\
                Denoising~(DAVIS)      & -                        & -                       & \bd{34.36} / 18.3 / \bd{220} & 33.45 / \bd{9.8} / \gf{131}         & \gf{35.18}/\gf{9.5}/433             \\
            \end{tabular}}
    \end{center}
\end{table}
\begin{figure*}[!h]
    \begin{center}
        \includegraphics[width=0.95\textwidth]{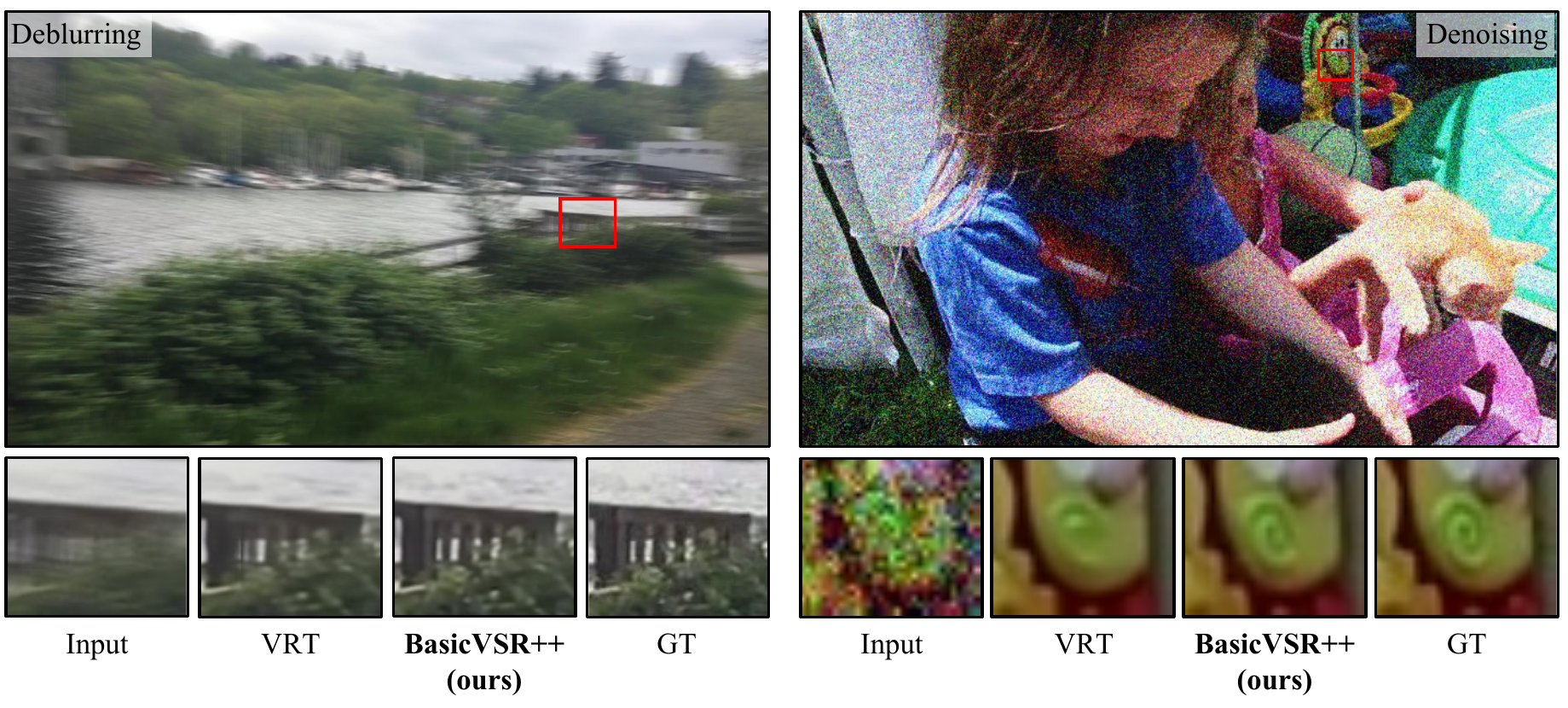}
        \caption{\textbf{Qualitative comparison with VRT. } Despite its smaller complexity, BasicVSR++$_{\downarrow4}$ is able to achieve comparable results to VRT.}
        \label{fig:qualitative}
    \end{center}
\end{figure*}

\subsection{Comparison to Transformer-Based Approaches}
Recently, Transformer-based methods~\cite{cao2021video,liang2022vrt,lin2022flow} have shown competitive performance in various video restoration tasks, including super-resolution, deblurring, and denoising. In this section, we compare BasicVSR++ (recurrent framework) to the Transformer-based methods. The comparison is shown in Table~\ref{tab:transformer}. 

Transformer-based methods achieve remarkable performance in the aforementioned tasks. For example, 
VRT~\cite{liang2022vrt} achieves comparable performance to BasicVSR++$_{\downarrow4}$ in deblurring and outperforms BasicVSR++$_{\downarrow4}$ in denoising. However, these methods generally require large network complexity to achieve good performance. For instance, for super-resolution, VSRT~\cite{cao2021video} and VRT~\cite{liang2022vrt} consist of over 30M parameters, which are about 5 times of BasicVSR++. In contrast, BasicVSR++ exploits long-term information through a recurrent network, achieving a promising performance-parameter-speed tradeoff. As shown in Fig.~\ref{fig:qualitative}, through exploiting long-term information, BasicVSR++ is able to reconstruct sharper edges and clearer details, achieving results highly similar to the ground truth. Furthermore, with a smaller downsampling factor, our BasicVSR++$_{\downarrow2}$ achieves a substantial performance gain compared to the aforementioned methods. 

\subsection{Performance-Speed Tradeoff with Input Downsampling}
As shown in our experiments, our BasicVSR++ framework is flexible in balancing speed and performance. On the one hand, better efficiency is obtained through shifting the computations to lower resolution. On the other hand, performance is substantially improved when more spatial information is preserved, with a slight sacrifice of efficiency. In practice, one could determine the optimal downsampling factor based on the task requirement.

%% file: main.bbl
\begin{thebibliography}{10}\itemsep=-1pt

\bibitem{arias2015towards}
Pablo Arias and Jean-Michel Morel.
\newblock Towards a bayesian video denoising method.
\newblock In {\em International Conference on Advanced Concepts for Intelligent
  Vision Systems}, 2015.

\bibitem{cao2021video}
Jiezhang Cao, Yawei Li, Kai Zhang, and Luc Van~Gool.
\newblock Video super-resolution transformer.
\newblock {\em arXiv preprint arXiv:2106.06847}, 2021.

\bibitem{chan2021basicvsr}
Kelvin~C.K. Chan, Xintao Wang, Ke Yu, Chao Dong, and Chen~Change Loy.
\newblock {BasicVSR}: The search for essential components in video
  super-resolution and beyond.
\newblock In {\em CVPR}, 2021.

\bibitem{chan2021understanding}
Kelvin~C.K. Chan, Xintao Wang, Ke Yu, Chao Dong, and Chen~Change Loy.
\newblock Understanding deformable alignment in video super-resolution.
\newblock In {\em AAAI}, 2021.

\bibitem{chan2021basicvsrpp}
Kelvin~C.K. Chan, Shangchen Zhou, Xiangyu Xu, and Chen~Change Loy.
\newblock {BasicVSR++}: Improving video super-resolution with enhanced
  propagation and alignment.
\newblock In {\em CVPR}, 2022.

\bibitem{charbonnier1994two}
Pierre Charbonnier, Laure Blanc-Feraud, Gilles Aubert, and Michel Barlaud.
\newblock Two deterministic half-quadratic regularization algorithms for
  computed imaging.
\newblock In {\em ICIP}, 1994.

\bibitem{mmedit}
MMEditing Contributors.
\newblock {MMEditing: OpenMMLab Image and Video Editing Toolbox}, 2022.

\bibitem{dai2017deformable}
Jifeng Dai, Haozhi Qi, Yuwen Xiong, Yi Li, Guodong Zhang, Han Hu, and Yichen
  Wei.
\newblock Deformable convolutional networks.
\newblock In {\em ICCV}, 2017.

\bibitem{davy2018non}
Axel Davy, Thibaud Ehret, and Gabriele Facciolo.
\newblock Non-local video denoising by {CNN}.
\newblock {\em arXiv preprint arXiv:1811.12758}, 2018.

\bibitem{haris2019recurrent}
Muhammad Haris, Greg Shakhnarovich, and Norimichi Ukita.
\newblock Recurrent back-projection network for video super-resolution.
\newblock In {\em CVPR}, 2019.

\bibitem{hyun2015generalized}
Tae Hyun~Kim and Kyoung Mu~Lee.
\newblock Generalized video deblurring for dynamic scenes.
\newblock In {\em CVPR}, 2015.

\bibitem{khoreva2018video}
Anna Khoreva, Anna Rohrbach, and Bernt Schiele.
\newblock Video object segmentation with language referring expressions.
\newblock In {\em ACCV}, 2018.

\bibitem{kingma2014adam}
Diederik Kingma and Jimmy Ba.
\newblock Adam: A method for stochastic optimization.
\newblock In {\em ICLR}, 2015.

\bibitem{lai2017deep}
Wei-Sheng Lai, Jia-Bin Huang, Narendra Ahuja, and Ming-Hsuan Yang.
\newblock Deep laplacian pyramid networks for fast and accurate
  super-resolution.
\newblock In {\em CVPR}, 2017.

\bibitem{li2021arvo}
Dongxu Li, Chenchen Xu, Kaihao Zhang, Xin Yu, Yiran Zhong, Wenqi Ren, Hanna
  Suominen, and Hongdong Li.
\newblock {ARVo}: Learning all-range volumetric correspondence for video
  deblurring.
\newblock In {\em CVPR}, 2021.

\bibitem{liang2022vrt}
Jingyun Liang, Jiezhang Cao, Yuchen Fan, Kai Zhang, Rakesh Ranjan, Yawei Li,
  Radu Timofte, and Luc Van~Gool.
\newblock {VRT}: A video restoration transformer.
\newblock {\em arXiv preprint arXiv:2201.12288}, 2022.

\bibitem{lin2022flow}
Jing Lin, Yuanhao Cai, Xiaowan Hu, Haoqian Wang, Youliang Yan, Xueyi Zou,
  Henghui Ding, Yulun Zhang, Radu Timofte, and Luc Van~Gool.
\newblock Flow-guided sparse transformer for video deblurring.
\newblock {\em arXiv preprint arXiv:2201.01893}, 2022.

\bibitem{loshchilov2016sgdr}
Ilya Loshchilov and Frank Hutter.
\newblock {SGDR}: Stochastic gradient descent with warm restarts.
\newblock In {\em ICLR}, 2017.

\bibitem{maggioni2011video}
Matteo Maggioni, Giacomo Boracchi, Alessandro Foi, and Karen Egiazarian.
\newblock Video denoising using separable {4D} nonlocal spatiotemporal
  transforms.
\newblock In {\em Image Processing: Algorithms and Systems}, 2011.

\bibitem{nah2017deep}
Seungjun Nah, Tae~Hyun Kim, and Kyoung~Mu Lee.
\newblock Deep multi-scale convolutional neural network for dynamic scene
  deblurring.
\newblock In {\em CVPR}, 2017.

\bibitem{nah2019recurrent}
Seungjun Nah, Sanghyun Son, and Kyoung~Mu Lee.
\newblock Recurrent neural networks with intra-frame iterations for video
  deblurring.
\newblock In {\em CVPR}, 2019.

\bibitem{pan2020cascaded}
Jinshan Pan, Haoran Bai, and Jinhui Tang.
\newblock Cascaded deep video deblurring using temporal sharpness prior.
\newblock In {\em CVPR}, 2020.

\bibitem{son2021ntire}
Sanghyun Son, Suyoung Lee, Seungjun Nah, Radu Timofte, Kyoung~Mu Lee,
  Kelvin~C.K. Chan, et~al.
\newblock {NTIRE 2021} challenge on video super-resolution.
\newblock In {\em CVPRW}, 2021.

\bibitem{su2017deep}
Shuochen Su, Mauricio Delbracio, Jue Wang, Guillermo Sapiro, Wolfgang Heidrich,
  and Oliver Wang.
\newblock Deep video deblurring for hand-held cameras.
\newblock In {\em CVPR}, 2017.

\bibitem{suin2021gated}
Maitreya Suin and A.~N. Rajagopalan.
\newblock Gated spatio-temporal attention-guided video deblurring.
\newblock In {\em CVPR}, 2021.

\bibitem{tao2018scale}
Xin Tao, Hongyun Gao, Xiaoyong Shen, Jue Wang, and Jiaya Jia.
\newblock Scale-recurrent network for deep image deblurring.
\newblock In {\em CVPR}, 2018.

\bibitem{tassano2019dvdnet}
Matias Tassano, Julie Delon, and Thomas Veit.
\newblock {DVDnet}: A fast network for deep video denoising.
\newblock In {\em ICIP}, 2019.

\bibitem{tassano2020fastdvdnet}
Matias Tassano, Julie Delon, and Thomas Veit.
\newblock {FastDVDnet}: Towards real-time deep video denoising without flow.
\newblock In {\em CVPR}, 2020.

\bibitem{vaksman2021patch}
Gregory Vaksman, Michael Elad, and Peyman Milanfar.
\newblock Patch craft: Video denoising by deep modeling and patch matching.
\newblock In {\em ICCV}, 2021.

\bibitem{wang2019edvr}
Xintao Wang, Kelvin~C.K. Chan, Ke Yu, Chao Dong, and Chen~Change Loy.
\newblock {EDVR}: Video restoration with enhanced deformable convolutional
  networks.
\newblock In {\em CVPRW}, 2019.

\bibitem{wieschollek2017learning}
Patrick Wieschollek, Michael Hirsch, Bernhard Scholkopf, and Hendrik Lensch.
\newblock Learning blind motion deblurring.
\newblock In {\em ICCV}, 2017.

\bibitem{xiang2020deep}
Xinguang Xiang, Hao Wei, and Jinshan Pan.
\newblock Deep video deblurring using sharpness features from exemplars.
\newblock {\em TIP}, 2020.

\bibitem{xue2019video}
Tianfan Xue, Baian Chen, Jiajun Wu, Donglai Wei, and William~T Freeman.
\newblock Video enhancement with task-oriented flow.
\newblock {\em IJCV}, 2019.

\bibitem{yang2021ntire}
Ren Yang, Radu Timofte, Jing Liu, Yi Xu, Xinjian Zhang, Minyi Zhao, Shuigeng
  Zhou, Kelvin~CK Chan, Shangchen Zhou, Xiangyu Xu, et~al.
\newblock {NTIRE} 2021 challenge on quality enhancement of compressed video:
  Methods and results.
\newblock In {\em CVPRW}, 2021.

\bibitem{zhang2018adversarial}
Kaihao Zhang, Wenhan Luo, Yiran Zhong, Lin Ma, Wei Liu, and Hongdong Li.
\newblock Adversarial spatio-temporal learning for video deblurring.
\newblock {\em TIP}, 2018.

\bibitem{zhou2019spatiotemporal}
Shangchen Zhou, Jiawei Zhang, Jinshan Pan, Haozhe Xie, Wangmeng Zuo, and Jimmy
  Ren.
\newblock Spatio-temporal filter adaptive network for video deblurring.
\newblock In {\em ICCV}, 2019.

\bibitem{zhu2019deformable}
Xizhou Zhu, Han Hu, Stephen Lin, and Jifeng Dai.
\newblock Deformable convnets v2: More deformable, better results.
\newblock In {\em CVPR}, 2019.

\end{thebibliography}
